# DIRECTED REDUCTION ALGORITHMS AND DECOMPOSABLE GRAPHS

**by**

| | | |
|---|---|---|
| **Ross D. Shachter,** | **Stig K. Andersen, and** | **Kim L. Poh** |
| Department of | Department of Medical Informatics | Department of |
| Engineering Economic Systems | and Image Analysis | Engineering Economic Systems |
| Stanford University | Aalborg University | Stanford University |
| Stanford, CA 94305-4025 | visiting the Department of | |
| shachter@sumex-aim.stanford.edu | Medical Informatics, Stanford University | |

## ABSTRACT

In recent years, there have been intense research efforts to develop efficient methods for probabilistic inference in probabilistic influence diagrams or belief networks. Many people have concluded that the best methods are those based on undirected graph structures, and that those methods are inherently superior to those based on node reduction operations on the influence diagram. We show here that these two approaches are essentially the same, since they are explicitly or implicity building and operating on the same underlying graphical structures. In this paper we examine those graphical structures and show how this insight can lead to an improved class of directed reduction methods.

## 1. Introduction

In recent years, there have been intense research efforts to develop efficient methods for probabilistic inference in probabilistic influence diagrams or belief networks. As these networks become increasingly popular representations for capturing uncertainty in expert systems, the performance of inference procedures is essential for normative reasoning in real time. To date, the best exact techniques for general probabilistic influence diagrams appear to those based on analogous undirected graphical structures [Andersen et al., 1989; Jensen et al., 1990a; Jensen et al., 1990b; Lauritzen and Spiegelhalter, 1988; Shafer and Shenoy, 1990]. Some people have also concluded that those methods are inherently superior to those based on node reduction operations on the influence diagram [Shachter, 1986; Shachter, 1988]. We show here that these two approaches are essentially the same, since they are explicitly or implicity building and operating on the same underlying graphical structures. In this paper we examine those graphical structures and show how this insight can lead to an improved class of directed reduction methods.

The key to this connection is the decomposable probabilistic influence diagram [Smith, 1989]. The main results in this paper are based on the connections Chyu has established between such diagrams and the undirected graph methods, allowing similar efficient computations using directed reduction operations [Chyu, 1990a; Chyu, 1990b]. By recognizing these connections and specialized reduction operations which exploit evidence nodes in the probabilistic influence diagram [Shachter, 1989], we can obtain complexity of the same order with both undirected and directed reduction methods.

In Sections 2 and 3 we introduce the notation and framework of the directed probabilistic influence diagram and undirected moral graph, respectively. Section 4 explains the use of the arc reversal operation to transform influence diagrams and Section 5 presents the corresponding operations to incorporate evidence into the diagram. These pieces are integrated into a new directed reduction method in Section 6, and some conclusions and extensions are presented in Section 7.

## 2. Probabilistic Influence Diagrams

A probabilistic influence diagram is a network built on a directed acyclic graph. The nodes in the diagram correspond to uncertain quantities, which can be observed, while the arcs indicate the conditioning relationships among those quantities. A decomposable probabilistic influence diagram is a special type of influence diagram whose properties will be explored throughout this paper.

A probabilistic influence diagram (PID) is a network structure built on a directed acyclic graph [Howard and Matheson, 1984]. Each node j in the set N= {1, ... ,n } corresponds to a random variable $X_j$. Each variable $X_j$ has a set of possible outcomes and a conditional probability distribution $\pi_j$ over those outcomes. The conditioning variables for $\pi_j$ have indices in the set of parents or conditional predecessors $C(j) \subset N$, and are indicated in the graph by arcs into node j from the nodes in C(j). Each variable $X_j$ is initially unobserved, but at some time its value $x_j$ might become known. At that point it becomes an evidence variable, its index is included in the set of evidence variables E, and this is represented in the diagram by drawing its node with shading.

As a convention, lower case letters represent single nodes in the graph and exact observations while upper case letters represent sets of nodes and random variables. If J is a set of nodes, $J \subseteq N$, then $X_J$ denotes the vector of variables indexed by J. For example, the conditioning variables for $X_j$ are denoted by $X_{C(j)}$ and might take on values $x_{C(j)}$.

In addition to the parents, we can define the children or (direct) successors of a node j. It is also convenient to keep track of the ancestors or indirect predecessors of node j which are defined to include the parents of j. Likewise, the ancestral set of node j is the ancestors of node j, plus j itself. Finally, the nondescendants ND(j) of node j are those nodes which are neither direct nor indirect successors of node j. ND(j) does not include node j itself.

Because there might be some observed evidence nodes in the diagram, some care must be taken to interpret the meaning of the distribution $\pi_j$ within node j. When there is no evidence then $\pi_j$ is simply the probability for $X_j$ given its conditioning variables, $P\{X_j \mid X_{C(j)}\}$. However, in general $\pi_j$ is defined as conditional on its nondescendant evidence nodes,

$$P\{ X_j \mid X_{C(j)}, X_{E \cap ND(j)} = x_{E \cap ND(j)} \}.$$

A PID will be called a decomposable PID (DPID) if there is an arc between every two nodes with a common child. It will be called decomposable with respect to node j if the subgraph induced by j's ancestral set is decomposable. It can be shown that if a PID is decomposable, then every subgraph of it is decomposable as well [Chyu, 1990b].

A list of the nodes N in a directed graph is said to be ordered if none of the parents of a node follow it in the list. Such a list exists if and only if there is no directed cycle among the nodes. Whenever a PID is decomposable with respect to a node j, there is a unique ordered list for the ancestral set of j [Chyu, 1990b].

One graph will be said to be consistent with another if both have the same nodes but the former has a subset of the arcs of the latter. A graph satisfying certain conditions is said to be minimal if there is no other graph consistent with it that satisfies those conditions.

## 3. Moral Graphs and Chordal Graphs

Moral and chordal graphs are undirected graph structures which correspond closely to PID's. The nodes have the same meanings, but there are many directed graphs corresponding to any undirected one. To appreciate the qualities of DPID's we need to explore the relationships between PID's and their undirected analogs.

Given a PID, its corresponding moral graph is obtained by adding undirected arcs between any nodes with a common child and dropping the directions from all of the arcs. For example, a PID corresponding to an example of incest in genetics [Jensen et al., 1990b] is shown in Figure 1a and its corresponding moral graph is shown in Figure 1b. The undirected arcs which were added between nodes with common children are drawn with dashed lines. Although the moral graph of a PID is unique, there can be many PID's corresponding to the same moral graph.

A moral graph is called a chordal graph if every cycle of four nodes or more possesses a chord, an arc between two nodes in the cycle which is not itself in the cycle. (Chordal graphs are also called triangulated [Berge, 1973; Golumbic, 1980] and decomposable.) A listing of the nodes in an undirected graph is said to be perfect if, for every node j in the list, there are arcs between all of the nodes which are adjacent to j and precede it in the list. A moral graph is chordal if and only if it has a perfect list [Golumbic, 1980]. For example, the graph shown in Figure 1d is a minimal chordal graph corresponding to the moral graph shown in Figure 1b. It is clearly chordal since ( B A C D F G H E I J ) is a perfect list. It is minimal because it would not be chordal without both of the added arcs, drawn with dashed lines. It is not unique, however, since there can be many minimal chordal graphs corresponding to the same moral graph. A perfect ordering can always be found, if one exists, by using <u>maximum cardinality search</u> [Tarjan and Yannakakis, 1984].

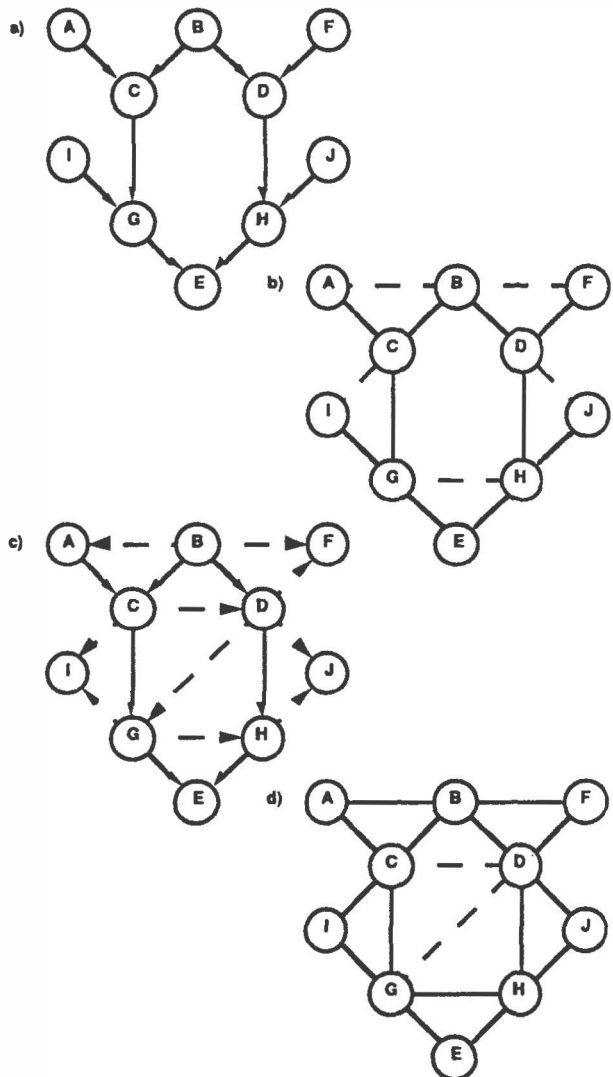

*Figure 1. Different graphical representations for the incest example.*

There is a strong relationship between DPID's and chordal graphs, stated in following theorem.

### Theorem 1.

A PID is decomposable if and only if its ordered list is perfect on its moral graph.

### Proof:

Given a DPID, its moral graph can be obtained without adding any new arcs. Because it is decomposable its ordered list will be perfect for that moral graph.

Given a perfect list on a moral graph, directions can be added to the undirected arcs from earlier nodes in the list to later ones. This will be an ordered list for the PID and the PID will be decomposable because the list was perfect.

We can always obtain a DPID from a chordal graph by using one of its perfect lists as an ordered list. Such a DPID is shown in Figure 1c using the perfect list ( B A C D F G H E I J ). The dashed arcs are the ones that were added or modified from the original PID shown in Figure 1a. Similarly, the moral graph for a DPID is that corresponding chordal graph, and any ordered list for the DPID will be perfect for the chordal graph. There is only one other result needed to characterize their relationship, defining the minimal DPID in terms of an original PID and a desired or <u>target</u> ordered list.

### Theorem 2.

Given a target ordered list and a moral graph, there corresponds a unique minimal DPID.

### Proof:

Starting with the last node in the list, add undirected arcs to the moral graph until the ordered list is perfect, so it is an ordered list for the corresponding DPID. There was no choice which arcs to add, and the list would not be perfect if any of the new arcs were not added, so the DPID is both unique and minimal.                                   #

Although there is no unique minimal chordal graph in general corresponding to a given moral graph, there is only one for which a target ordered list is perfect. As a result, we can summarize the relationships between a PID and its associated DPID's, moral graphs, and chordal graphs. Given a PID there is a unique moral graph. Given that moral graph and a target ordered list, there is a unique minimal DPID. Finally, the moral graph for the DPID is the minimal chordal graph corresponding to the original PID for which the target ordering is perfect.

## 4. Influence Diagram Transformations

The arc reversal operation transforms one PID into another with a different ordered list. In the process, extra arcs often must be added to the PID. However, in transforming to and from DPID's, we can guarantee limits on the addition of those extra arcs.

The arc reversal operation transforms a PID by changing the direction of one of the arcs [Olmsted, 1983; Shachter, 1986]. Afterwards, each of the two nodes inherits their common parents. The operation can be interpreted as momentarily merging the two nodes and then splitting them apart. The arc (i, j) is reversible if it is the only directed path from i to j. Otherwise, a directed cycle would be created by reversing the arc. The general case for arc reversal is shown in Figure 2, in which the arc (i, j) is reversed. Afterwards, A, B, and C are parents for both i and j.

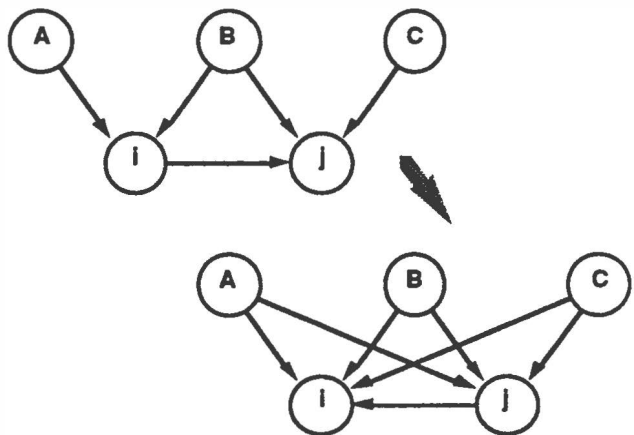

*Figure 2. General Arc Reversal Operation.*

Given a target ordered list and any PID, we can transform the PID into another PID consistent with the minimal DPID, using only arc reversal operations [Chyu, 1990b; Shachter, 1990]. The algorithm involves visiting each node j in the reverse target order: reverse all arcs to j's successors which come before it in the target order. Arcs must be reversed in the order they appear in the current PID, but when there is a choice, reverse in the target order. This algorithm creates the minimal number of additional arcs. Its correctness is given by the following theorem.

### Theorem 3. Transforming to Target DPID

Given a PID and a target ordered list, a new PID consistent with the corresponding minimal DPID can be obtained through a sequence of arc reversals.

### Proof:

The proof is by induction as we visit each node k in the reverse target order. We have two induction hypotheses: the current PID contains no arcs outside of the target DPID and the target list after k is an ordered list for the current PID.

To prove the theorem we must show that all of the arcs reversed are reversible, and that the induction hypotheses are maintained.

First, we show that the target list starting with k will be ordered for the current PID. This follows, because we reverse any arcs from k to successors which precede it in the ordered list, and all of the other nodes which follow k are already in the target order.

Second, we show that any arc (k, j) to be reversed is indeed reversible. If this were not true, then there must be some node i, k → i → j. If i belongs before k then (k, i) would have been reversed before (k, j). Therefore i must belong after k, but it is not properly ordered since it precedes j. Thus we contradict the induction hypotheses and it must be true that arc (k, j) is reversible.

Finally, we must show that no arcs are created outside of the target DPID. A new arc is created when we reverse arc (k, j) only if there is some node i≠k which is a parent of k or j and not of the other. Since all nodes following k are in their target order, k must follow both i and j in the target order. Now all current arcs are by induction in the target chordal graph, so this new arc is required for the target ordering to be perfect. #

As a special case of this result, we can transform between any two DPID's which have the same moral graph, and hence correspond to the same chordal graph [Chyu, 1990b; Smith, 1989].

### Corollary 1.

We can transform one DPID to another DPID corresponding to the same chordal graph through a sequence of arc reversals.

By similar reasoning, it can be shown that any arc reversal operations on a DPID will keep it decomposable [Chyu, 1990b]. However, the resulting PID will not in general be a minimal DPID for the starting PID and the target order.

### Theorem 4.

If a sequence of arc reversal operations are performed on a DPID, it will continue to be decomposable.

## 5. Evidence Transformations

New observations are incorporated into a PID in two steps. First, the evidence is absorbed into the network and then it is propagated throughout the network using the evidence reversal operation, a variant of arc reversal. In the process, new arcs are added until the PID eventually becomes a DPID. At all times, the PID represents the posterior joint distribution.

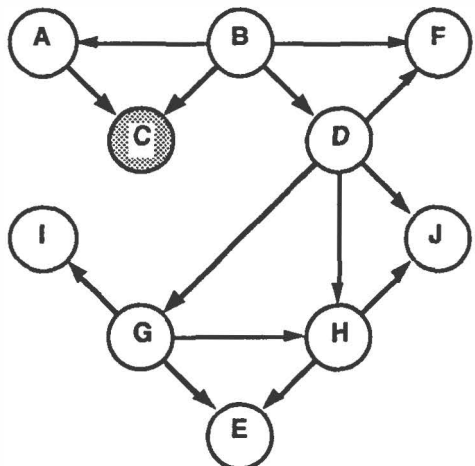

*Figure 3. DPID after evidence absorption of C.*

The operation of evidence absorption maintains the posterior joint distribution while recognizing the observation of an exact value for a variable in the PID [Lauritzen and Spiegelhalter, 1988; Shachter, 1989]. There is no longer any need to maintain distributions with the other possible outcomes for the variable in its node or in the nodes of its children. Therefore, when $X_j$ is observed at value $x_j$, the conditional distribution $\pi_j$ becomes a likelihood function, $P\{ X_j = x_j \mid X_{C(j)} \}$, and the arc to each child k of j is absorbed, since the conditional distribution $\pi_k$, $P\{ X_k \mid X_{C(k)}, X_j = x_j \}$, no longer depends on $X_j$. For example, if C were observed in the DPID shown in Figure 1c, then after evidence absorption we obtain the DPID shown in Figure 3.

The operation of evidence absorption does not destroy the decomposability of a PID, since all of the arcs from the observed node to its children are absorbed.

### Proposition 1.

If evidence absorption is performed on a DPID, it remains decomposable.

When evidence is absorbed at a node the distributions of its ancestors are affected indirectly. To propagate these effects throughout the network, we must reorder the PID so that the evidence node has no ancestors. This reordering process consists of a sequence of specialized reversal operations. Evidence reversal of the arc (i, j) is closely related to arc reversal, except that because the successor node is observed there is no need for it to have a child afterward [Shachter, 1989]. The general case is shown in Figure 4. It can be thought of as arc reversal followed by evidence absorption, but it is more efficient to recognize the special properties of evidence reversal.

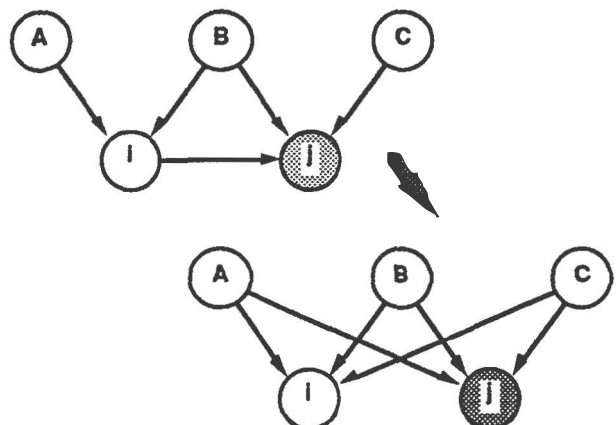

*Figure 4. General Evidence Reversal Operation.*

When evidence reversal is performed on arc (i, j), evidence node j moves one step closer to the start of an ordered list for the PID. In order for it to have no ancestors, the operation will have to be performed on each ancestor in reverse order. This sequence of evidence reversal operations is called evidence propagation [Lauritzen and Spiegelhalter, 1988;

Shachter, 1989]. Afterwards, node j will have neither parents nor children. For example, consider the part of the incest PID shown in Figure 5a. We have some evidence about C, but it is not an exact observation of C, so we create a variable K whose exact observation describes the evidence for C. (Node K is created and absorbed at the same time, so it has no children and its distribution is simply a likelihood function for C. Because K has no children and only one parent, a DPID would remain decomposable after it was added.) Evidence propagation consists of evidence reversals with C, A, and B in turn until node K is disconnected as shown in Figure 5b, 5c, and 5d. There was a choice whether to reverse A before B since they were not ordered beforehand in the PID. Notice that they are ordered afterward and the PID has become a DPID in the process, as will be proven in general below. Finally, the evidence could have originally related to multiple nodes as in shown in Figure 5e. This works best when the nodes are a subset of a clique [Golumbic, 1980]. Each node and its parents are contained in some clique.

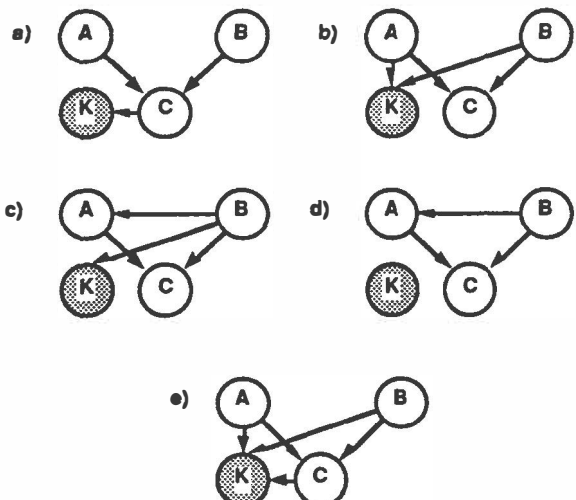

*Figure 5. Application of evidence propagation on part of the incest example.*

Theorem 5.
  Once evidence propagation has been performed from node j in a PID, the PID will be decomposable with respect to node j.
Proof:
  Consider any node i with multiple parents in the ancestral set for j, and let l and m be any two of those parents. Both l and m are parents of the evidence node after the arc from i is reversed. Without loss of generality, suppose the arc from l to the evidence node is reversed before the arc from m. Afterwards m will be a parent of l.                    #

By this same logic, if the ancestral set is already decomposable then no new arcs will be created by evidence propagation. Also, since the ancestral set will be decomposable, and the ancestral set of the sink nodes (nodes without children) is the entire PID, evidence at all of the sink nodes will result in a DPID.

Corollary 2.
  When evidence propagation is performed on a DPID, no new arcs are created.

Corollary 3.
  Once evidence propagation has been performed from all sink nodes in a PID, the PID will be decomposable.

Evidence propagation can be performed efficiently even when evidence has been absorbed at multiple nodes in the PID. Each unobserved node in the network has to be visited once, in reverse graph order: if it has no evidence children, there is nothing to do; if it has exactly one, then perform evidence reversal; otherwise it must have multiple evidence children, and they can be combined into one evidence child by multiplying their likelihood functions so that a single evidence reversal can be performed. (If some of the multiple evidence children have multiple parents, then the resulting product has all of their parents.)

In summary, the operations of evidence absorption and evidence propagation eventually result in a DPID. If the PID is already decomposable and evidence is only within cliques, then those operations will never add new arcs.

## 6. Putting it All Together

In this section, we assemble the results from throughout the paper to develop a directed reduction algorithm to compute the posterior joint distribution. Because the choice of chordal graph is arbitrary, we can obtain precisely the same chordal graph as in the best undirected methods [Andersen et al., 1989; Jensen et al., 1990a; Jensen et al., 1990b; Lauritzen and Spiegelhalter, 1988; Shafer and Shenoy, 1990] with similar complexity using directed reduction operations.

The first step in this process is to determine a target ordered list for the DPID. The list can either be selected directly or, if a chordal graph is chosen instead, one of its perfect lists should be used. One way to generate the perfect list is to perform maximum cardinality search on the chordal graph, using an ordered list for the original PID to break ties [Chyu, 1990b; Tarjan and Yannakakis, 1984].

Using this list and the algorithm described in Section 4, we can pre-reverse arcs to obtain a PID consistent with the unique minimal DPID. For example, given target ordered list ( B A C D F G H E I J ), a perfect list for the chordal graph is shown in Figure 1d. We can pre-reverse arcs from the original PID shown in Figure 1a to obtain the PID shown in Figure 6. The shaded, dashed arcs would not appear in this PID, but we can infer them from the target ordering. If they were present, we would have the DPID shown in Figure 1c.

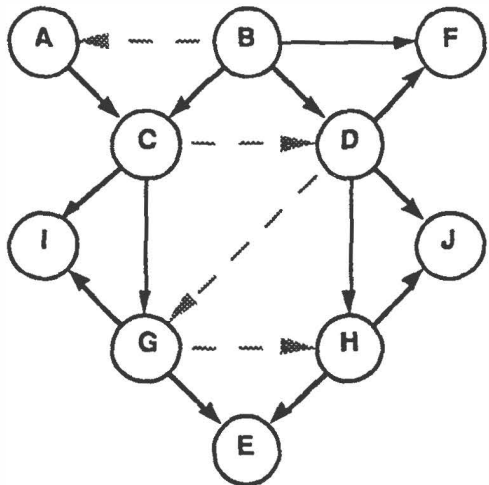

*Figure 6. PID for evidence propagation after pre-reversals.*

We can now perform evidence absorption and propagation on the PID. In the process, the shaded, dashed arcs in Figure 6 might have to be added. In the worst case, they will all appear and we will obtain the target DPID. If the evidence absorption is exact evidence about nodes in the network, then those nodes and their incident arcs will be absorbed through evidence absorption and propagation. If the evidence is about nodes in the same clique, then that evidence can be absorbed and propagated while maintaining a PID consistent with the target DPID. If there are multiple observations, then evidence propagation should be performed in reverse order throughout the PID to avoid duplicate operations.

In this method, we maintain an updated posterior joint distribution for the PID given the evidence. If we desire any general conditional distributions, they can be obtained through reduction operations [Shachter, 1988]. If we want posterior marginal distributions for the variables in the PID, they can be obtained by a probability propagation process [Lauritzen and Spiegelhalter, 1988; Shachter, 1989] operating on the cliques. By comparing the basic operations performed by the different methods, we can verify that they have the same order of complexity. This is because at each step they are performing similar tasks on the same graphical structure and operating on data structures of the same maximal dimensions. There can, of course, be significant differences in the actual computation times.

Proposition 2.
    The directed reduction method and the undirected methods of HUGIN and Lauritzen-Spiegelhalter are of the same order of complexity.

## 7. Conclusions and Extensions

We have shown that a directed reduction algorithm can perform operations on the same graph and of the same order of complexity as the best undirected methods for probabilistic inference. This result can be interpreted in two ways. First, pre-reversals allow us to use the best possible choice of chordal graph so we can learn from the undirected methods a superior strategy for reduction. Second, we can plainly see how the chordal graph structure represents the "worst case" for posterior joint dependence. No matter what evidence (within cliques) is observed, no additional arcs will be necessary to represent the posterior PID.

Some natural extensions to the directed algorithm are to exploit efficiencies which have been developed in either the directed or undirected representations.

In the directed representation, an important property is that of a deterministic function, a variable whose outcome is known with certainty given its parents' outcomes. This introduces additional conditional independence into the diagram which can be exploited during evidence propagation. At the same, when the PID is only being used to obtain posterior marginal distributions for a subset of variables or with limited observations, then the PID can be preprocessed to eliminate variables that are irrelevant for the desired results [Geiger et al., 1989; Shachter, 1988; Shachter, 1990]. This elimination can be performed on the working PID or, if possible, before the target DPID is determined.

Another promising hybrid might exploit the impressive speed and simplicity of the HUGIN undirected method [Andersen et al., 1989; Jensen et al., 1990a; Jensen et al., 1990b] by maintaining joint distributions for a node and its parents instead of conditional distributions. This simplifies the operation of arc reversal, but does require maintaining the full DPID instead of simply a PID consistent with it. There are a couple of advantages to using this method on undirected graphs which appear applicable to directed methods as well. These advantages are symmetric operations for evidence and probability propagation and the recognition of zeros in the sparse joint distribution matrices.

## 8. Acknowledgements

We are grateful for the comments and suggestions of Richard Barlow, Stephen Chyu, and Robert Fung.